\documentclass[runningheads]{llncs}
\usepackage{graphicx}

\usepackage{amsmath}         
\usepackage{graphicx}        
\usepackage{accents}         
\usepackage{multirow}        
\usepackage{mathtools}       
\usepackage{footnote}
\usepackage{tablefootnote}
\usepackage{esvect}          
\usepackage{tabularx}        
\usepackage{subfiles}
\usepackage{colortbl}        
\usepackage{lipsum}          
\usepackage{xcolor}
\usepackage{caption}         
\usepackage{wrapfig}         

\usepackage[misc]{ifsym}
\usepackage{booktabs}
\usepackage{amsfonts}
\usepackage{subfig}
\usepackage{tabularx}  
\usepackage{hyperref}

\makeatletter
\DeclareRobustCommand{\cev}[1]{%
  {\mathpalette\do@cev{#1}}%
}
\newcommand{\do@cev}[2]{%
  \vbox{\offinterlineskip
    \sbox\z@{$\m@th#1 x$}%
    \ialign{##\cr
      \hidewidth\reflectbox{$\m@th#1\vec{}\mkern4mu$}\hidewidth\cr
      \noalign{\kern-\ht\z@}
      $\m@th#1#2$\cr
    }%
  }%
}
\makeatother

\newcommand{\savefootnote}[2]{\footnote{\label{#1}#2}}
\newcommand{\repeatfootnote}[1]{\textsuperscript{\ref{#1}}}

\newcommand{\repeatthanks}{\textsuperscript{\thefootnote}}

\begin{document}
%
\title{PU GNN: Chargeback Fraud Detection in P2E MMORPGs via Graph Attention Networks with Imbalanced PU Labels}

\titlerunning{PU GNN for Chargeback Fraud Detection}

\toctitle{PU GNN: Chargeback Fraud Detection in P2E MMORPGs via Graph Attention Networks with Imbalanced PU Labels}

\tocauthor{Jiho~Choi,Junghoon~Park,Woocheol~Kim,Jin-Hyeok~Park,Yumin~Suh,Minchang~Sung}

%
%




\author{
    Jiho Choi\thanks{Both authors contributed equally to this work.}\orcidID{0000-0002-7140-7962}
    \and Junghoon Park\repeatthanks\orcidID{0000-0002-6054-9113}
    \and Woocheol Kim
    \and Jin-Hyeok Park
    \and Yumin Suh
    \and Minchang Sung ({\Letter})
}


\institute{
    Netmarble Corp., Republic of Korea \\
    \email{\{jihochoi, jhoonpark, kwc4616, realhyeok, yuum0131, sungmirr\}@netmarble.com} \\
} 
\authorrunning{Choi and Park}



%



%
\maketitle              

\begin{abstract}



The recent advent of play-to-earn (P2E) systems in massively multiplayer online role-playing games (MMORPGs) has made in-game goods interchangeable with real-world values more than ever before. The goods in the P2E MMORPGs can be directly exchanged with cryptocurrencies such as Bitcoin, Ethereum, or Klaytn via blockchain networks. Unlike traditional in-game goods, once they had been written to the blockchains, P2E goods cannot be restored by the game operation teams even with chargeback fraud such as payment fraud, cancellation, or refund. To tackle the problem, we propose a novel chargeback fraud prediction method, \textit{PU GNN}, which leverages graph attention networks with PU loss to capture both the players' in-game behavior with P2E token transaction patterns. With the adoption of modified GraphSMOTE, the proposed model handles the imbalanced distribution of labels in chargeback fraud datasets. The conducted experiments on three real-world P2E MMORPG datasets demonstrate that \textit{PU GNN} achieves superior performances over previously suggested methods.




\end{abstract}

\section{Introduction}


\begin{figure}
    \centering \includegraphics[width=0.8\textwidth]{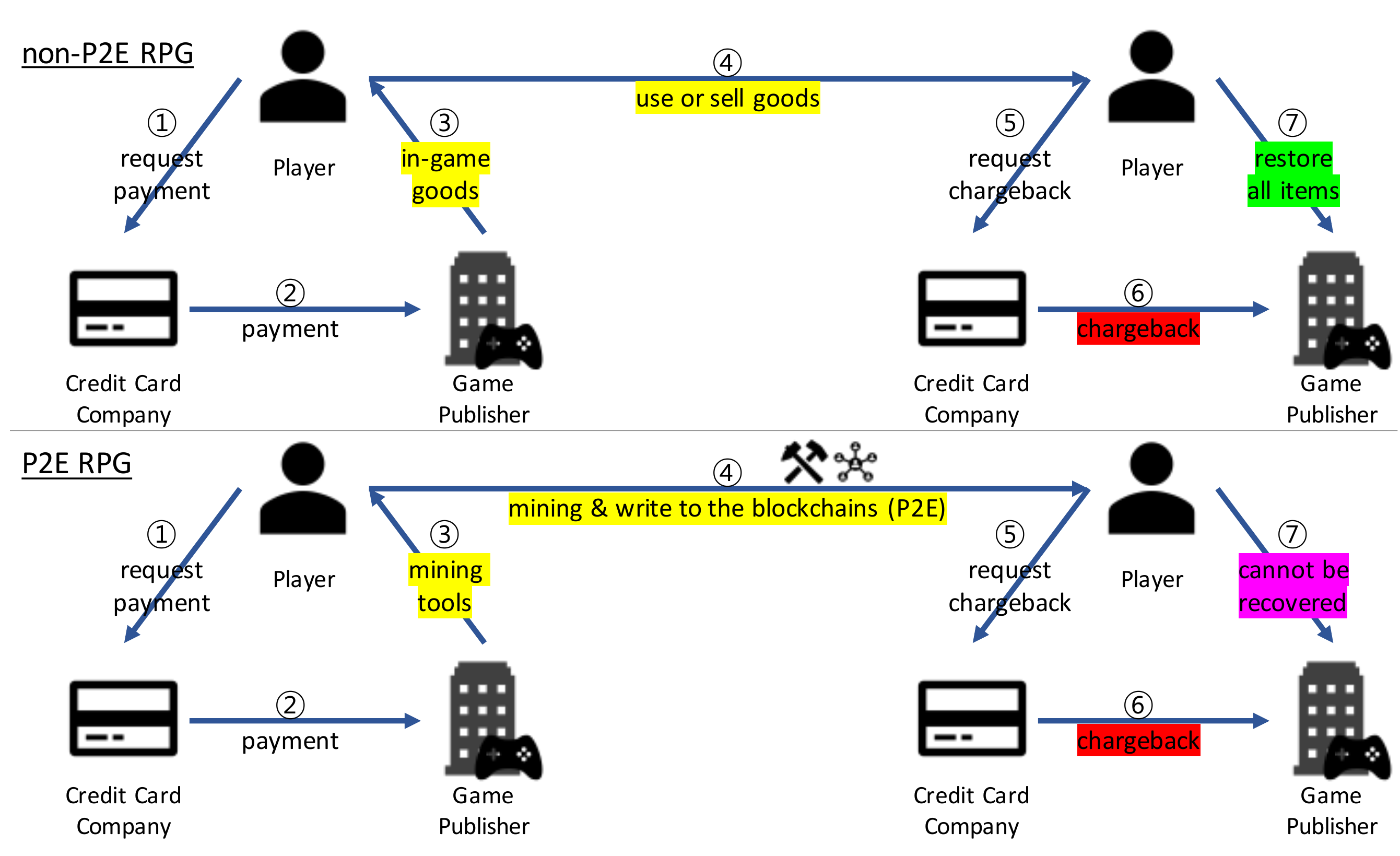}
    \caption{
        The brief comparisons of goods refunds in non-P2E MMORPGs and P2E MMORPGs.
        \textmd{
            Unlike non-P2E goods, it is challenging for game publishers to roll back the P2E goods even in the case of payment cancellation.
        }
    }
    \label{figure_01}
\end{figure}


The recent advent of play-to-earn (P2E) systems, a new paradigm where game players can earn real-world value through their in-game activities, in massively multiplayer online role-playing games (MMORPGs) has made the value of in-game goods interchangeable with real-world values \cite{lee2018no,kiong2022metaverse} more than ever before. The goods in the P2E MMORPGs can be directly exchanged for cryptocurrencies such as Bitcoin, Ethereum, and Klaytn via blockchain networks, immersing players to put more endeavors and facilitating players for better engagements. However, unlike traditional in-game goods, from the game operation teams' or publishers' perspective, P2E goods are vulnerable to chargeback fraud \cite{guo2018sell,liu2022cfledger} such as payment fraud, cancellation, or refund abuses. The P2E tokens or comparable P2E goods cannot be restored by the game operation teams once it had been written to the blockchain. Figure~\ref{figure_01} is a brief comparison of game goods refunds of non-P2E and P2E games. In general, goods obtained through players' payments are subject to retrieval in the event of payment cancellation. However, the P2E goods or tokens are difficult to retrieve as they are recorded on the blockchain even if the payments are canceled. Thus it is essential to detect chargeback frauds in P2E MMORPGs.


Some of the previously conducted studies \cite{carneiro2017data,de2018customized,mittal2019performance} had utilized financial datasets with traditional machine learning methods such as naive Bayes, random forest, logistic regression, and K-means clustering for chargeback fraud detection. \cite{carneiro2017data,de2018customized} leveraged the learning methods with under-sampling to overcome the nature of label imbalance between fraud and benign. \cite{jurgovsky2018sequence,rao2021xfraud} adopted sequence and transaction modeling methods to combat chargeback fraud and fraud transactions. Although these approaches had revealed some of the patterns in chargeback fraud, they are not quite fitted with chargeback fraud in P2E MMORPGs since they had the limitation of not jointly considering both the player behaviors and token transaction patterns. Chargeback frauds in P2E MMORPG often occur in automated programs rather than manual playing by individuals. Therefore, the fraudsters leave relatively distinct in-game behavior action patterns such as action sequences and intervals. The details will be discussed in Section \ref{ssec:player_behavior_modeling} and Section \ref{ssec:graph_attention_networks}. In addition, players who have not yet canceled their payment should be considered unlabeled rather than fixed as negative since they could cancel their payment after finishing mining the P2E token with their paid in-game goods.



Recent advents in graph neural networks (GNNs) have shown promising results of graph inference tasks in many different domains such as social media, biochemistry, knowledge graphs, citation networks, and transaction graphs \cite{kipf_17,hamilton2017inductive,velivckovic2018graph,Xu_19,brody2021attentive,gong2019exploiting,wang2019heterogeneous,peng2020spatial}. Likewise, positive-unlabeled (PU) learning has been adopted in various fields of study. Some studies in bioinformatics \cite{li2022positive,yang2018adasampling} use PU learning to overcome the lack of labeling, and \cite{elkan2008learning,scott2009novelty,wu2021learning} utilized PU learning for anomaly detection and outlier detection.


In this study, we overcome the previously addressed problem in chargeback fraud detection in P2E MMORPGs with the recent advent of GNNs and PU learning. We propose a novel chargeback fraud detection method, \textit{PU GNN}. The proposed method utilizes players' in-game activity logs and P2E token transaction histories with positive and unlabeled label settings. The performance evaluation on three real-world datasets demonstrated the model's superiority over other previously presented methods.

We summarized our contributions as follows:

\begin{enumerate}
    \item We propose a novel chargeback fraud detection method, \textit{PU GNN}, for play-to-earn massively multiplayer online role-playing games (P2E MMORPGs).
    \item The proposed method carefully utilizes both the players' in-game activities with P2E token transactions and tackled label imbalance with an over-sampling method and positive \& unlabeled label setups.
    \item The conducted experiments on three real-world datasets demonstrate the method is superior to previously presented methods.
\end{enumerate}

\section{Related Work}


%

\subsection{Fraud Detection}



Fraud detection has been studied in various fields with the perspectives of  credit card fraud, payment fraud, and online game fraud \cite{kou2004survey,carneiro2017data,carcillo2021combining,chen2021deep,makki2019experimental}. Traditional rule-based approaches \cite{kou2004survey} had been extended to pattern-based learning methods \cite{carneiro2017data,de2018customized} by discovering the distinctiveness patterns between fraudsters and benign users. \cite{mittal2019performance,carcillo2021combining,chen2021deep} adopted unsupervised approaches to retrieve outlier scores to spot fraudulent activities, and \cite{makki2019experimental} had utilized an imbalanced setup by leveraging one-class classification and nearest neighbors approach. Recently, \cite{jurgovsky2018sequence,yu2018netwalk,rao2021xfraud} had adopted sequential or graph structures in transaction logs or graph representation to spot anomaly interactions.


\subsection{Graph Neural Networks}
Graph neural networks (GNNs) have shown promising results in many different node, link, and graph inference tasks \cite{kipf_17,hamilton2017inductive,velivckovic2018graph,Xu_19,brody2021attentive}.
Some of the GNNs variants leverage more on specific setups such as edge representation \cite{gong2019exploiting}, heterogeneous node types \cite{wang2019heterogeneous}, or dynamic graphs \cite{peng2020spatial}. However, one of the key mechanisms that are shared with GNNs is message passing (or neighborhood aggregation) which takes the topologically connected components into account when learning the representation of an entity. A brief generalization of GNNs is as follows:
\begin{equation}
    \label{equation_generalized_GNN}
    h_{i}^{(k)} = \underbrace{
        \mathtt{UPDATE}_{\theta} \biggl(
            h_{i}^{(k-1)},
            \overbrace{\mathtt{AGG}_{\theta} \Bigl(h_{j}^{(k-1)}, \forall j \in \mathcal{N}_{i} \Bigl)}^{\text{aggregating neighbors' representations}}
        \biggl)
    }_{\text{updating current node's representation}}
\end{equation}
The hidden representation of $i^{th}$ node after passing through ${k}^{th}$ GNN layer, $h_i^{(k)}$, can be retrieved with the combination of $\mathtt{AGG}_{\theta}$ and $\mathtt{UPDATE}_{\theta}$. A differentiable (learnable) and permutation invariant function $\mathtt{AGG}_{\theta}$ \cite{hamilton2017inductive} aggregates (sometimes sampled) set of neighborhood node $\mathcal {N}{(\cdot)}$. $\mathtt{UPDATE}_{\theta}$ is an injective update function \cite{Xu_19} to associate current states $h_{i}^{(k-1)}$ with aggregated neighbors'. Some of the recent variants of GNNs such as graph attention networks \cite{velivckovic2018graph,brody2021attentive} adopt attention mechanism \cite{vaswani2017attention}, and assign different weights when combining the neighbor nodes' embeddings.

\subsection{Imbalanced Positive \& Unlabeled Learning}

In the classification task, it is important for the labeled data to be evenly distributed, or otherwise, the classifiers could overfit the majority classes \cite{lemaitre2017imbalanced,mani2003knn,wilson1972asymptotic}. To handle the problem, sampling studies such as under-sampling or over-sampling have been actively conducted. Under-sampling methods such as \cite{mani2003knn} solved the problem of data imbalance by eliminating the dominant class by finding data points that do not belong to its K nearest data point labels and balancing the label distribution with other classes. An over-sampling method such as SMOTE \cite{chawla2002smote} duplicates the minority class observations by interpolating the nearest data points in the same class.

In addition, especially in industrial fields, the lack of labeled data with the majority of data being unlabeled, some studies \cite{lee2003learning,elkan2008learning,christoffel2016class,du2014analysis} use unlabeled data through a method called positive-unlabeled (PU) learning. The method can help the classifier by leveraging unlabeled data. The method regards unlabeled as negative labels with the mixture of positive, or estimating the approximated risks \cite{du2015convex,du2014analysis}. PU learning has been widely adopted in various studies of classification, anomaly detection, and outlier detection. \cite{nguyen2011positive,liu2002partially,zhou2021pure,wu2021learning}.


\section{Problem Definition}


Let $\mathcal{G}=(\mathcal{V},\mathcal{E})$ be the \textit{P2E token transaction graph} where the vertex set $\mathcal{V}=\{p_1, p_2, ..., p_{|\mathcal{V}|}\}$ and edge set $\mathcal{E}=\{t_1, t_2, ..., t_{|\mathcal{E}|}\}$ denote ${|\mathcal{V}|}$ \underline{p}layers and ${|\mathcal{E}|}$ token \underline{t}ransfers respectively. The node $p$ consists of \textit{in-game behavioral} features $x \in \mathcal{R}^{F_x}$, and a link $t$ includes \textit{token transfer} features $e \in \mathcal{R}^{F_e}$. A node belongs to one of the binary class $y \in \{-1, 1\}$, and $p(x, y)$ is the joint density of $(\mathcal{X}, \mathcal{Y})$. $p_p(x)=p(x|Y=+1)$ and $p_n(x)=p(x|Y=-1)$ are positive and negative marginals respectively with $p(x)$ being the whole $\mathcal{X}$ marginal.

The aim is to learn a decision function $f$: $\mathcal{R}^{F_x} \rightarrow \mathcal{R}$ to predict \textit{chargeback fraud} with positive and unlabeled (PU) labels. We formulate the task as a node classification task in a token transaction graph $\mathcal{G}$ with a PU learning setup in which each set of data points is sampled from $p_p (x)$, $p(x)$, which are $\mathcal{X}_p=\{x_i^p\}_{i=1}^{n_p} \sim p_p(x)$ and $X_u=\{x_i^u\}_{i=1}^{n_u} \sim p(x)$.

\section{Proposed Method}

\begin{figure*}
    \centering \includegraphics[width=1\textwidth]{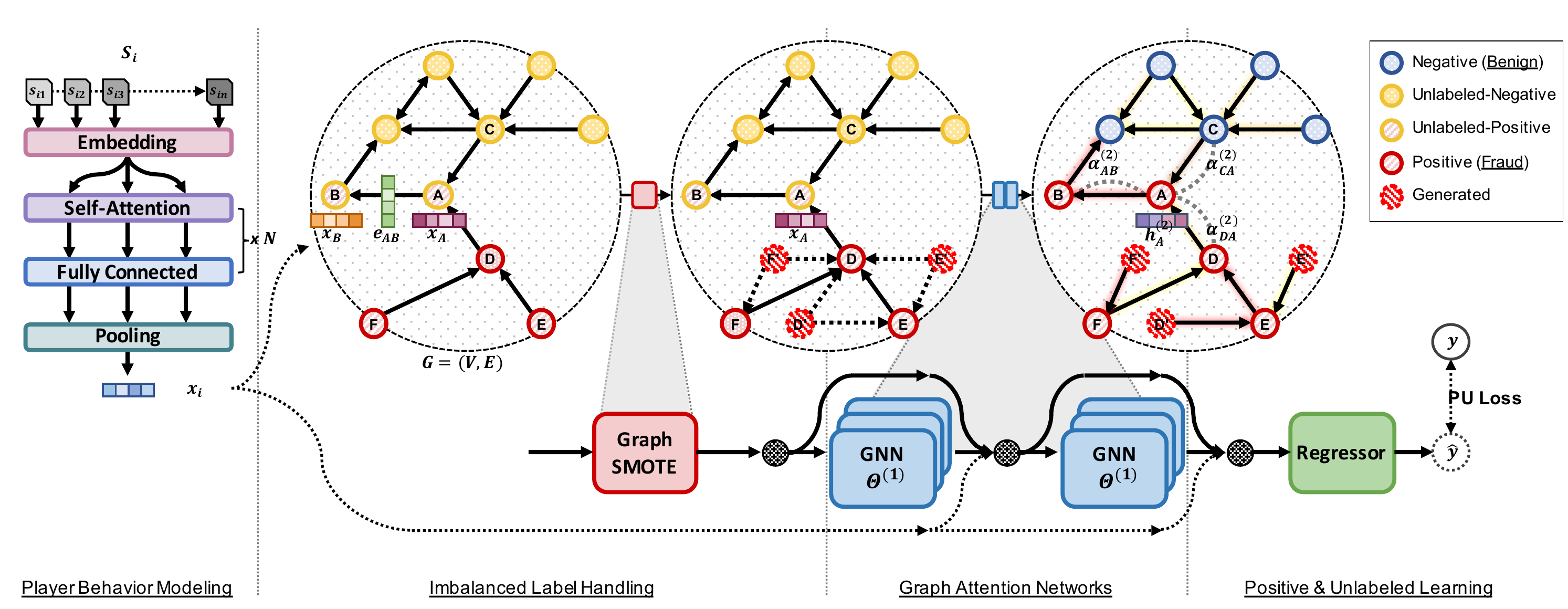}
    \caption{
        The overall architecture of \textit{PU GNN} for chargeback fraud detection in P2E MMORPG.
        \textmd{
            The proposed method consists of four main components. The behavior modeling layer finds representations of the initial node features by learning behavior sequences. There are two different labels; unlabeled (a mixture of fraud ($\mathcal{A, B}$) and benign ($\mathcal{C}$)) and fraud ($\mathcal{D, E, F}$), with the respect to transaction networks. Notice that nodes with fewer labels ($\mathcal{D, E, F}$) are augmented with GraphSMOTE \cite{zhao2021graphsmote} and balanced. The graph attention layers \cite{brody2021attentive} calculate the attention weights of node $\mathcal{A}$ (expressed as edge colors) to better leverage the embedding of the neighbor nodes $\mathcal{B, C, D}$. With the PU learning setup, half of the positive nodes are treated as unlabeled.
        }
    }
    \label{figure_model}
\end{figure*}

In this section, we introduce \textit{PU GNN} architecture as Figure~\ref{figure_model}, which includes player behavior modeling with in-game activity logs, graph attention with P2E token transaction graph, and calculating positive and unlabeled loss with imbalanced label distribution.

\subsection{Player Behavior Modeling}
\label{ssec:player_behavior_modeling}

%


\begin{figure}
    \centering \includegraphics[width=0.8\textwidth]{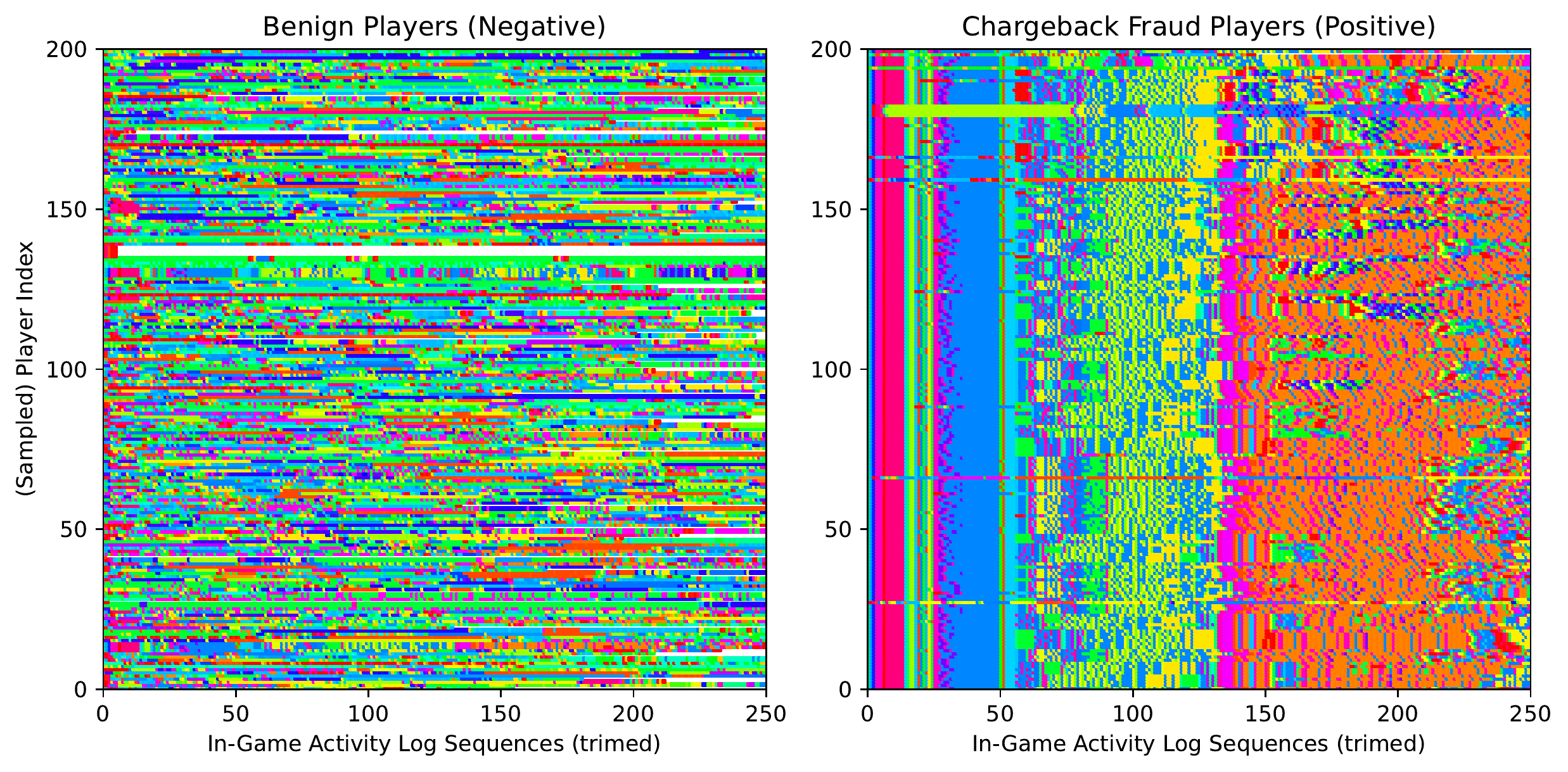}
    \caption{
        The sampled users' activity log sequences of benign and chargeback fraud players in P2E MMORPGs.
        \textmd{
            The chargeback fraud users (or automated programs) show collusive activities which mainly focus on purchasing P2E goods and mining tokens. (each color denotes different types of activity logs; login, dungeon entrance, guild join, quest completion, mining, et cetera)
        }
    }
    \label{figure_02_activity_logs}
\end{figure}


We first retrieve the player $i$'s initial behavioral representation ${x_i}$ from his $n$ most recent \textit{in-game activity logs} ${S_i} = \{s_{i1},s_{i2}, \cdots, s_{in}\}$ such as character login, item purchase, item use, skill use, quest completion, level up, goods acquisition, dungeon entrance, guild join, and so on (may vary by game).
Figure~\ref{figure_02_activity_logs} is the  brief comparison of benign and fraud players where chargeback fraud users (or automated programs such as a macro) produce collusive activities which largely focus the mining the P2E goods or preceding activities such as the completion of tutorials or quests. With a simple embedding lookup layer, $s_{i(\cdot)}$ are mapped to $d$ dimensional vectors resulting $ {S'_i} \in \mathcal{R}^{n \times d}$. To better spot future chargeback players, not all relations of activity logs should be considered equally. The scaled dot-product attention \cite{vaswani2017attention,kenton2019bert} with a fully connected layer, a common sequence modeling method, is adopted as below to utilize the activity logs:
\begin{equation}
    \label{equation_Transformer}
    {H_i} = FC\{\mathtt{Attention} (Q, K, V)\} = FC\{softmax(\frac{Q K^T}{\sqrt{d}}) V\}
\end{equation}, where $query; Q$, $key; K$, and $value; V$ are all ${S'_i}$. The attended behavioral logs ${H_i}$ are then $\mathtt{mean}$ and $\mathtt{max}$ pooled to model the player's in-game activity. The concatenation, $\oplus$, of pooled results is the initial behavioral representation as follows:
\begin{equation}
    \label{equation_h_i}
    {x_i} = \{ \mathtt{Pool}_\mathtt{MEAN}({H_i}) \oplus  \mathtt{Pool}_\mathtt{MAX}({H_i}) \} \in \mathcal{R}^{F_x}
\end{equation}
, and now it is treated as the node feature of player $i$.

\subsection{Graph Attention Networks}
\label{ssec:graph_attention_networks}

To better capture the relations between players and chargeback attempts, \textit{play-to-earn (P2E) token transaction graph} $\mathcal{G}=(\mathcal{V},\mathcal{E})$ is utilized. The transfer records in the transaction graph may include the high-dimensional hidden relations of source and target nodes \cite{liu2019hyperbolic,shen2021identity} such as ownership, collusive works, and so on. We adopt and modify graph attention networks (GATv2) \cite{brody2021attentive} where dynamic attention is computed. The previously retrieved node features $x \in \mathcal{R}^{F_x}$ and link features $e \in \mathcal{R}^{F_e}$ are first upsampled with GraphSMOTE \cite{zhao2021graphsmote} and the details would be discussed in Section~\ref{section_4_3}.

We leverage all features of the source node, target node, and link to calculate the attention coefficients. A shared attention mechanism $ \mathtt{a} : \mathcal{R}^{F_x} \times \mathcal{R}^{F_x} \times  \mathcal{R}^{F_e} \rightarrow \mathcal{R}$ is as below to calculate the attention weight $\alpha_{ij}$:
\begin{equation}
    \alpha_{ij}^{(l)}  = \frac{
        exp\bigr( \mathtt{a}^{(l)}  \mathtt{LeakyReLU} ( \Theta ^ {(l)} \bigr[ x_{i} \mathbin \Vert x_{j} \mathbin \Vert e_{ij} \bigr] ) \bigr)
    }{
        \sum_{
            k \in \mathcal{N}_{i} \cup {\{i\}}
        } \{
            exp\bigr( \mathtt{a}^{(l)} \mathtt{LeakyReLU} ( \Theta ^ {(l)} \bigr[ x_{i} \mathbin \Vert x_{k} \mathbin \Vert e_{ik} \bigr] ) \bigr) \}  
    }
\end{equation}, where $\mathtt{a}^{(l)}$ and $\Theta ^ {(l)}$ are learnable parameters for ${l}^{th}$ GNN layer. With the calculated attention coefficients with generalized message passing framework Equation~\ref{equation_generalized_GNN}, the neighborhood nodes' representations are aggregated and combined as:

\begin{equation}
    \vec{h}_{i}^{(l)} = \sum_{j \in \mathcal{N}_{i} \cup {\{i\}}} \alpha_{ij}^{(l)} \Theta^{(l)} \vec{h}_{i}^{(l-1)}
\end{equation}, where $h_i^{(0)} = x_i$. With two GATv2 layers, the node representation embedded the 1-hop and 2-hop neighbor nodes. The bi-directional $(h_{i} = \cev{h}_{i} \oplus \vec{h}_{i})$ and skip-connection \cite{tong2017image} are concatenated to better embed the node features. The model leverages the concatenation representation of $h_{i}$ with ${x_i}$ from Section~\ref{ssec:player_behavior_modeling} to classify the frauds.

\subsection{Imbalanced Positive \& Unlabeled Learning}
\label{section_4_3}

Non-negative PU (nnPU) \cite{kiryo2017positive} learning improved former PU learning, which is also known as unbiased PU (uPU) learning \cite{du2015convex}. uPU learning uses unbiased risk estimators. Let $\mathcal{L}$ : $R \times \{ \pm 1 \} \rightarrow R$ be the loss function. $\mathcal{L}(t,y)$ represents the loss while predicting an output $t$ and the ground truth $y$ and $f$ represents decision function. Denoting $R_{p}^{+} (f) = E_{X \sim p_p(x)}[\mathcal{L}((f(X), +1)]$ , ${R}_n^- (f)\ =\ E_{X \sim p_n(x)}[\mathcal{L}((f(X), -1)]$, ordinary binary classification risk estimator is directly approximated by:
\begin{equation}
    \widehat{R}_{pn}(f) = \pi_{p}\widehat{R}^+_p(f) + \pi_{n}\widehat{R}_n^-(f)
\end{equation}

Meanwhile, the PU learning setting has no information on negative data, but the risk estimator $R(f)$ can be approximated directly \cite{du2014analysis}. Denotes  $R_p^-(f)\ =\ E_p[\mathcal{L}((f(X),\ -1)]$ and $R_u^-(f) = E_{X\sim p(x)}[\mathcal{L}((f(X), -1)]$, then as $\pi_{n}p_n(x) = p(x) - \pi_{p}p_p(x)$, we can obtain 
\begin{equation}
    \pi_n R^-_n(x) = R_u^-(f) - \pi_{p}R^-_p(f)
\end{equation}

Using above, the unbiased risk of $R(f)$ is approximated directly by:
\begin{equation}
\begin{gathered}
    \widehat{R}_{pu}(f) =  \pi_p \widehat{R}_p^+(f) -\pi_p\widehat{R}_p^-(f)+\widehat{R}_u^-(f) \\
    \widehat{R}_p^-(f) = {\frac{1}{n_p}} \sum_{i=1}^{n_p} \mathcal{L}(f(x_i^p), -1),
    \widehat{R}_u^-(f) = {\frac{1}{n_u}}\sum_{i=1}^{n_u} \mathcal{L}(f(x_i^u),\ -1)
\end{gathered}
\end{equation}

To overcome the issues with the convergence rate and complex estimation error bounds, \cite{kiryo2017positive} suggested the non-negative risk estimator, which is denoted by:
\begin{equation}
    \widetilde{R}_{pu}(f) =\pi_p \widehat{R}_p^+(f) + max( 0,\widehat{R}_u^-(f)-\widehat{R}_p^-(f) )
\end{equation}

GraphSMOTE \cite{zhao2021graphsmote} oversampled minority class by introducing SMOTE \cite{chawla2002smote} in graph-structured data with training link generation simultaneously. The chargeback fraud data of P2E MMORPG have the characteristics of imbalance in the graph structure and the possibility of being labeled as a potential fraud even though it is not yet labeled. Therefore, we adopted GraphSMOTE \cite{zhao2021graphsmote} and nnPU learning \cite{kiryo2017positive} to handle the imbalance and the unlabeled situation in the dataset.

\subsection{~Loss, Training \& Inference}

\label{section_4_4}
In this study, we train our model with the empirical estimation of risk with the following loss function:
\begin{equation}
    \label{loss_in_nnpu}
    \mathcal{L} = \widetilde{R}_{pu}(f) = \pi_p \widehat{R}_p^+(f) + max( 0,\widehat{R}_u^-(f)-\widehat{R}_p^-(f) )
\end{equation}, with the sigmoid function, ${\ell}_{\texttt{sigmoid}}(t, y) = 1 / (1 + exp(t*y))$ for $\mathcal{L}(t,y)$ in $\widehat{R}_p^+(f)$, $\widehat{R}_u^-(f)$, $\widehat{R}_p^-(f)$ as previously described in Section~\ref{section_4_3}. To minimize $\mathcal{L}$ in equation \ref{loss_in_nnpu}, output $t$ has to be closer to label $y$.
The sigmoid function is adopted for $\mathcal{L}(t, y)$ since it is continuously differentiable across its entire domain and can be minimized by the gradient-based algorithms \cite{kiryo2017positive}. The training task of the loss function can be seen as a regression task since the output value, $\hat{y}$, of the proposed model, lies in $\hat{y} \in [-1, +1]$. For the ablation study to verify the effectiveness of the loss, the softmax function is used to retrieve $\hat{y}$, and the cross-entropy between $\hat{y}$ and one-hot label ${y}$ is used as loss as follows:
$\mathcal{L} = CE(y, \hat{y}) = -\sum_{i=1} \{ y_i log(\hat{y}_i) + (1-y_i)log(1-\hat{y}_i) \}$.


\section{Experiments}




This section introduces our datasets, baselines, implementation detail, experimental results, and ablation study. Three real-world datasets had been collected for the evaluation. Since the proposed method consists of player behavior modeling and a graph structure-based model, we focus on comparing ours with other approaches that can consider player behaviors or transaction histories.


\begin{figure}
    \centering

    \noindent\makebox[\textwidth]{\includegraphics[width=0.45\textwidth]{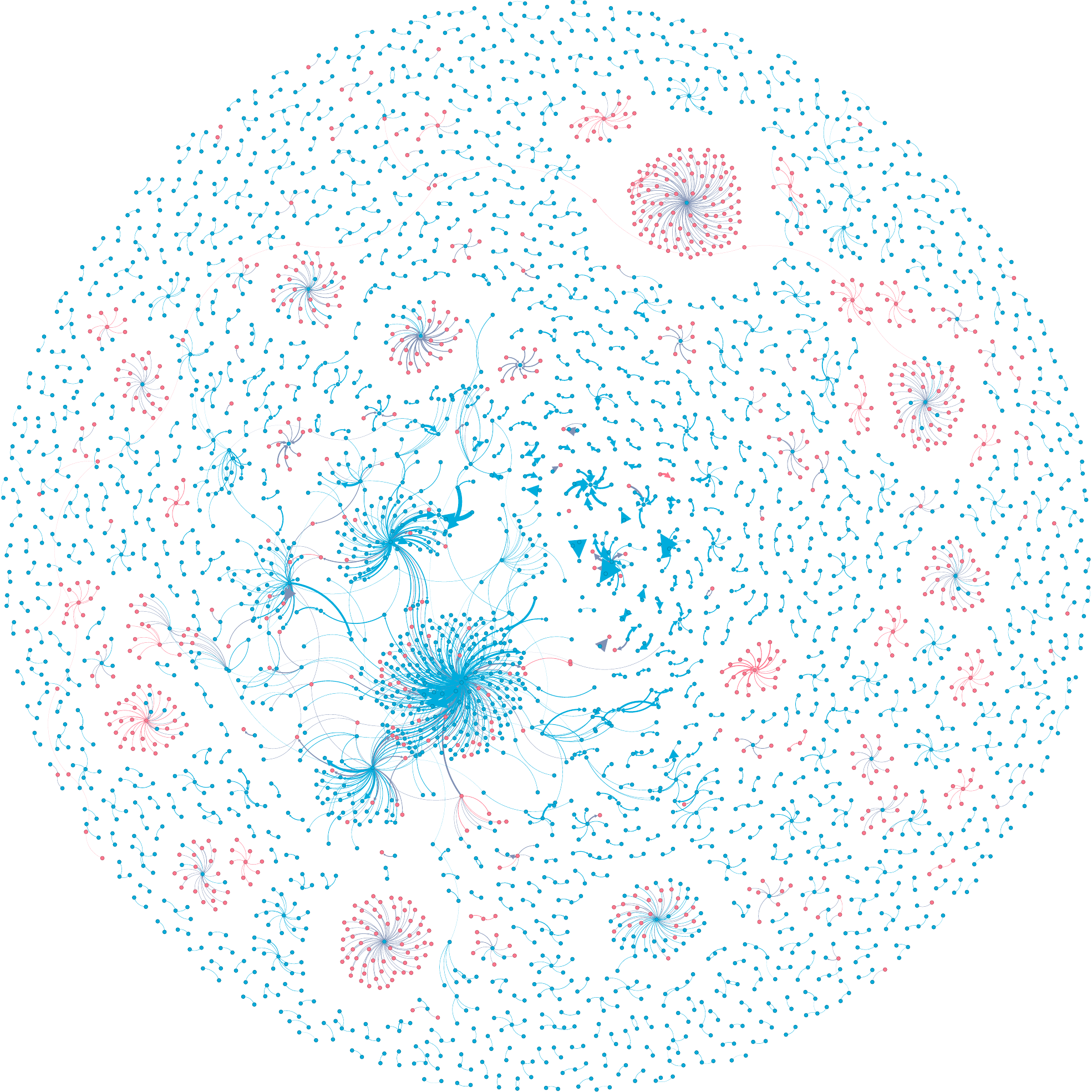}}
    
    \caption{
        A brief overview of sampled P2E token transaction networks. Each node and link represents a player in P2E MMORPG and a token transfer respectively. The colors of the nodes represent fraudsters (red) and unlabeled, a mixture of fraudsters and benign, users (blue).
    }
    \label{figure_00_network}
\end{figure}

\subsection{Experimental Setup}

\subsubsection{Datasets}




We retrieved three datasets from two popular P2E MMORPGs played globally; {$\texttt{MMORPG}_{01}$} and {$\texttt{MMORPG}_{02}$} (The titles of the games and token names have been anonymized). The datasets include players' behavior activity logs and P2E token transaction graphs. Figure~\ref{figure_00_network} is the brief overview of the sampled P2E token transactions. We chronologically and randomly sampled train, validation, and test sets. The chargeback fraud labels include manual annotation by game operation teams, with the consideration of the players' network and device information, and whether the players cancel the payments after transferring the purchased tokens. For the PU learning environment, we used half of the positive labeled observations as positive and the rest are treated as unlabeled. The test evaluations are done with positive and negative ground-truth labels. The detailed statistics of the datasets are described in Table~\ref{table_dataset}.

\begin{table}
    \caption{Statistical details of chargeback \& P2E MMORPG transaction graph datasets}
    \label{table_dataset}
    \centering
    \begin{tabular}{c r l r l r l}
        \toprule
        -                      & \multicolumn{2}{c}{$\texttt{DATASET}_{01}$}                      & \multicolumn{2}{c}{$\texttt{DATASET}_{02}$} & \multicolumn{2}{c}{$\texttt{DATASET}_{03}$} \\
        \midrule
        Game           & \multicolumn{2}{c}{$\texttt{MMORPG}_{01}$}                         & \multicolumn{2}{c}{$\texttt{MMORPG}_{01}$} & \multicolumn{2}{c}{$\texttt{MMORPG}_{02}$} \\
        P2E Token              & \multicolumn{2}{c}{{$\texttt{TOKEN}_{01}$}}                & \multicolumn{2}{c}{{$\texttt{TOKEN}_{01} + \texttt{TOKEN}_{02}$}} & \multicolumn{2}{c}{{$\texttt{TOKEN}_{03}$}} \\
        Date                   & \multicolumn{2}{c}{Jul.$\sim$Aug. 2022}                  & \multicolumn{2}{c}{Aug.$\sim$Oct. 2022}  & \multicolumn{2}{c}{May$\sim$Oct. 2022} \\
        

        $|\mathcal{V}|$        & \multicolumn{2}{c}{32.9K}                                  & \multicolumn{2}{c}{28.1K} & \multicolumn{2}{c}{5.6K} \\
        $|\mathcal{E}|$        & \multicolumn{2}{c}{62.4K}                                  & \multicolumn{2}{c}{67.4K} & \multicolumn{2}{c}{33.8K} \\
        Train ($P$:$N$)        & \quad \small{2.4K} & \small{: 10.5K}                       & \multicolumn{2}{c}{3.2K:9.8K} & \multicolumn{2}{c}{0.5K:2.3K} \\
        Validation             & \small{2.7K} & \small{:  8.1K}                             & \multicolumn{2}{c}{1.4K:4.2K}   & \multicolumn{2}{c}{0.2K:1.1K} \\
        Test                   & \small{2.1K} & \small{:  7.2K}                             & \multicolumn{2}{c}{2.3K:6.8K}  & \multicolumn{2}{c}{0.4K:1.1K} \\
        \bottomrule
    \end{tabular}
\end{table}


\subsubsection{Baselines}


We compared the proposed model with several baseline models; MLP, GRU \cite{chung2014empirical}, Self-attention \cite{vaswani2017attention}, GCN \cite{kipf_17}, GraphSAGE \cite{hamilton2017inductive}, GAT \cite{velivckovic2018graph}, OCGNN\savefootnote{footnote-label}{We implemented the simplified versions of the models which borrowed key ideas to adopt different situations from the original tasks.} \cite{wang2021one}, and DAGNN\repeatfootnote{footnote-label} \cite{li2022dual}. All non-PU experiments are done with cross-entropy loss and experiments on GNNs are done in an inductive setting \cite{hamilton2017inductive}.

%


We compare the proposed PU GNN model with the following fraud detection baseline models:
\begin{itemize}
\item MLP: Three-layer multi-layer perceptron classifier to map in-game behavior features to fraud or benign labels.
\item GRU \cite{chung2014empirical}: An simple staked-GRU model to learn players' sequential in-game behavior patterns with their labels.
\item Self-attention \cite{vaswani2017attention}: An self-attention mechanism to learn players' in-game behavior patterns.
\item GCN \cite{kipf_17}: A GNN-based model that learns contextual relations between senders and receivers in P2E transaction graphs.
\item GraphSAGE \cite{hamilton2017inductive}: A GNN-based model aggregating neighbors' embeddings with current representation to earn relations between senders and receivers in P2E transaction graphs.
\item GAT \cite{velivckovic2018graph} A GNN-based model with an attention mechanism to retrieve edge weights in transaction networks.
\item OCGNN \cite{wang2021one}: A GNN-based anomaly detection model extended OCSVM \cite{scholkopf1999support} to the graph domain to utilize hypersphere learning.
\item DAGNN \cite{li2022dual}: A GNN-based fraud detection model which leverages different augmented views of graphs called disparity and similarity augments.



\end{itemize}

\subsubsection{Implementation Detail}

The proposed model is built with PyTorch, PyTorch Geometric \cite{pytorch_geometric}, and BigQuery in the Google Cloud Platform. 5 self-attention with fully connected layer blocks and 2 GATv2 layer blocks are adopted. The embedding size, denoted as $F_x$ of \ref{equation_h_i}, is 128 dimensions. BatchNorm \cite{ioffe2015batch} and Dropout \cite{srivastava2014dropout} are added between layers and Adam optimizer \cite{kingma2014adam} and early stopping \cite{prechelt1998early} with the patience of 10 were adopted. We report the average results of 5 runs.


\subsection{Performance Evaluation}





\begin{table}
\centering
\caption{
    The performance evaluations of chargeback fraud detection in P2E MMORPGs. The evaluated metrics are F1-score, AUC, Pos (true positive rate; sensitivity), and Neg (true negative rate; specificity). The best results of the F1-score and AUC are in the underlines.
}
\label{table_performance}



\resizebox{1\textwidth}{!}
{
    \begin{tabular}{llcccccccccccc}
        \hline
        \multicolumn{2}{r}{~}            & \multicolumn{4}{c}{$\texttt{DATASET}_{01}$}         & \multicolumn{4}{c}{$\texttt{DATASET}_{02}$}         & \multicolumn{4}{c}{$\texttt{DATASET}_{03}$} \\
    
        \cmidrule(lr){3-6}
        \cmidrule(lr){7-10}
        \cmidrule(lr){11-14}


        \multicolumn{2}{l}{\multirow{2}{*}{Method}}       & \multirow{2}{*}{F1}  & \multirow{2}{*}{AUC}  & Pos   & Neg   & \multirow{2}{*}{F1}  & \multirow{2}{*}{AUC} & Pos   & Neg & \multirow{2}{*}{F1}  & \multirow{2}{*}{AUC} & Pos   & Neg \\
        \multicolumn{2}{l}{}                                       &       &       & \scriptsize{TPR}   & \scriptsize{TNR}      &      &      & \scriptsize{TPR}   & \scriptsize{TNR} &      &      & \scriptsize{TPR}   & \scriptsize{TNR} \\
        \cmidrule(lr){1-14}
        \multicolumn{2}{l}{MLP}                                         & \footnotesize{0.778}              & \footnotesize{0.665}             & \footnotesize{0.391}  & \footnotesize{0.940}        & \footnotesize{0.771}	& \footnotesize{0.762}	& \footnotesize{0.548}	& \footnotesize{0.976} & \footnotesize{0.775}&\footnotesize{0.798} &\footnotesize{0.698} &\footnotesize{0.897} \\
        \multicolumn{2}{l}{GRU \cite{chung2014empirical}}               & \footnotesize{0.811}              & \footnotesize{0.753}             & \footnotesize{0.634}  & \footnotesize{0.872}   & \footnotesize{0.847}	& \footnotesize{0.835}	& \footnotesize{0.706}	& \footnotesize{0.964} &\footnotesize{0.780} &\footnotesize{0.748} &\footnotesize{0.885} & \footnotesize{0.610}\\
        \multicolumn{2}{l}{Self-attention \cite{vaswani2017attention}}  & \footnotesize{0.805}              & \footnotesize{0.775}             & \footnotesize{0.725}  & \footnotesize{0.825}   & \footnotesize{0.831}	& \footnotesize{0.829}	& \footnotesize{0.814}	& \footnotesize{0.843} & \footnotesize{0.784}              & \footnotesize{0.768}             & \footnotesize{0.828}  & \footnotesize{0.707} \\
        \cmidrule(lr){1-14}
        \multicolumn{2}{l}{GCN \cite{kipf_17}}                          & \footnotesize{0.816}              & \footnotesize{0.734}             & \footnotesize{0.549}  & \footnotesize{0.919}   & \footnotesize{0.849}	& \footnotesize{0.846}	& \footnotesize{0.819}	& \footnotesize{0.872} & \footnotesize{0.776}              & \footnotesize{0.798}             & \footnotesize{0.703}  & \footnotesize{0.893} \\
        \multicolumn{2}{l}{GraphSAGE \cite{hamilton2017inductive}}      & \footnotesize{0.834}              & \footnotesize{0.791}             & \footnotesize{0.708}  & \footnotesize{0.875}        & \footnotesize{0.870}	& \footnotesize{0.863}	& \footnotesize{0.776}	& \footnotesize{0.950} & \footnotesize{0.791}              & \footnotesize{0.781}             & \footnotesize{0.813}  & \footnotesize{0.748} \\
        \multicolumn{2}{l}{GAT \cite{velivckovic2018graph}}             & \footnotesize{0.836}              & \footnotesize{0.847}             & \footnotesize{0.886}  & \footnotesize{0.807}        & \footnotesize{0.867}	& \footnotesize{0.859}	& \footnotesize{0.769}	& \footnotesize{0.948} & \footnotesize{0.782}              & \footnotesize{0.803}             & \footnotesize{0.710}  & \footnotesize{0.895} \\
    
        
        \multicolumn{2}{l}{OCGNN \cite{wang2021one}}             & \footnotesize{0.801}              & \footnotesize{0.763}             & \footnotesize{0.402}  & \footnotesize{0.844}        & \footnotesize{0.862}	& \footnotesize{0.852}	& \footnotesize{0.727}	& \footnotesize{0.988} & \footnotesize{0.781}              & \footnotesize{0.783}             & \footnotesize{0.703}  & \footnotesize{0.802} \\
        \multicolumn{2}{l}{DAGNN \cite{li2022dual}}             & \footnotesize{0.846}              & \footnotesize{0.835}             & \footnotesize{0.827}  & \footnotesize{0.831}        & \footnotesize{0.875}	& \footnotesize{0.869}	& \footnotesize{0.792}	& \footnotesize{0.946} & \footnotesize{0.801}              & \footnotesize{0.787}             & \footnotesize{0.812}  & \footnotesize{0.810} \\
        

        \cmidrule(lr){1-14}

        \multicolumn{2}{l}{PU GNN (Proposed)}             & \footnotesize{\underline{0.855}} & \footnotesize{\underline{0.854}} & \footnotesize{0.866} & \footnotesize{0.843}        & \footnotesize{\underline{0.884}}	& \footnotesize{\underline{0.876}}	& \footnotesize{0.788}	& \footnotesize{0.965} &\footnotesize{\underline{0.811}}  &\footnotesize{\underline{0.801}} &\footnotesize{0.832} & \footnotesize{0.771} \\

    
        \hline 
        
    \end{tabular}
}

\end{table}







\subsubsection{Overall Detection Results (F1, AUC, Recall, Specificity)}



%

We report the F1-score, ROC AUC, recall (true positive rate), and specificity (true negative rate) of the proposed model with baselines. The overall statistical details of performance evaluations are shown in Table~\ref{table_performance}. Methods such as MLP and GRU show that the chargeback fraud label could be classified with the aid of in-game behavioral features in all three datasets. Sequential models such as GRU and self-attention improved their performance by taking the sequential patterns of the behavior logs into account. The graph-based models such as GCN, GraphSAGE, and GAT outperform the sequence-based baselines; GRU, and self-attention. It is been demonstrated the necessity of using both the in-game behavior features and the token transfer network structures to detect chargeback fraud in P2E MMORPGs. However, the features are not the only thing that affects the performance. Although both types of behavior patterns and transaction patterns are utilized, the classification result of OCGNN shows that the one-class abnormal detection is not suitable for the chargeback fraud detection tasks. We believe that the task that we are tackling is out of the scope of anomaly and more fitted with classification. DAGNN utilizes augmented views of graphs and improves performance compares to other GNN-based models. The proposed method, \textit{PU GNN}, outperforms other baseline methods by leveraging PU learning which takes the assumption of un-chargeback paid players as unlabeled observations. Handling the label imbalance with modified GraphSMOTE helps the model to robustly learn player representations.






\subsection{Ablation Study}

\subsubsection{(Ablation Study 1) t-SNE visualization of the components' output embedding}

\begin{figure}
    \centering \includegraphics[width=1.0\textwidth]{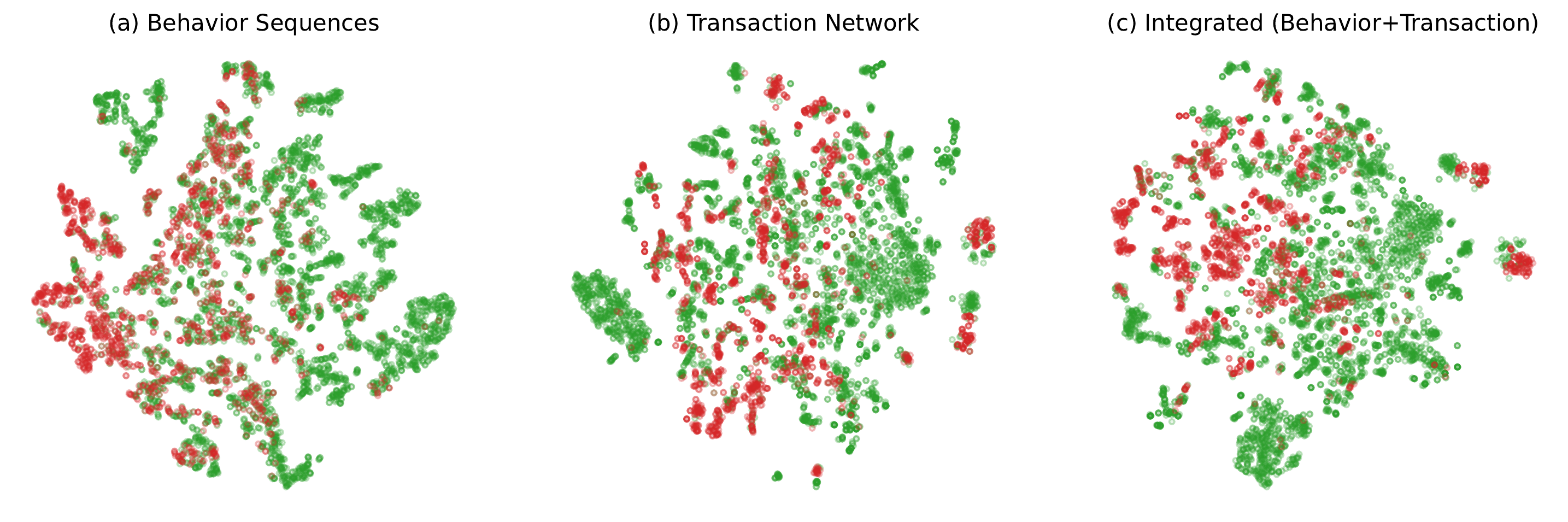}
    \caption{
        The t-SNE visualization of the sampled users' by each component of the proposed model. Each component handles (a) activity log sequences, (b) transaction patterns, and (c) both. The color represents benign (green) and chargeback fraud players (red) in P2E MMORPGs.
    }
    \label{figure_04_tsne}
\end{figure}


To take a closer look at what roles and advantages the components of our method have, we separated and compared the proposed method into three parts. Three components are (a) player behavior sequence modeling (Section~\ref{ssec:player_behavior_modeling}), (b) graph attention network (Section~\ref{ssec:graph_attention_networks}), and (c) leveraging both of them. Figure~\ref{figure_04_tsne} is the 2-dimensional t-SNE visualization of the output embedding of each of the three parts. First, the player behavior sequence modeling component utilizes players' behavior features. In Figure~\ref{figure_04_tsne}(a), large-scale clusters could be identified which represent similar behavior log patterns left between chargeback fraudsters. However, there were many overlapping parts with benign users and fraudsters, thus the features are not enough to correctly detect the frauds. Figure~\ref{figure_04_tsne}(b), where only the transaction networks are used as input features, shows the multiple clusters of small node counts. The result is led by connected nodes sharing closer embedding spaces. Finally, when both the behavior patterns and the transaction networks are used in Figure~\ref{figure_04_tsne}(c), there are advantages in fraud detection by leveraging the information of both the large and small clusters in previous components.


\subsubsection{(Ablation Study 2) Effects of PU and SMOTE}

We conducted another ablation study with different aspects to see the effectiveness of PU and SMOTE in the proposed method. The series of ablation studies are combinations of the below elements:
\begin{itemize}
\item {\bfseries w/o SMOTE}: Constituting the original input data and its preprocessing, training without the oversampling by GraphSMOTE \cite{zhao2021graphsmote}.
\item {\bfseries w/o PU}: Without PU loss for the circumstances of unlabeled, trains model with cross-entropy classification loss.
\end{itemize}

\begin{figure}
    \centering \includegraphics[width=1.0\textwidth]{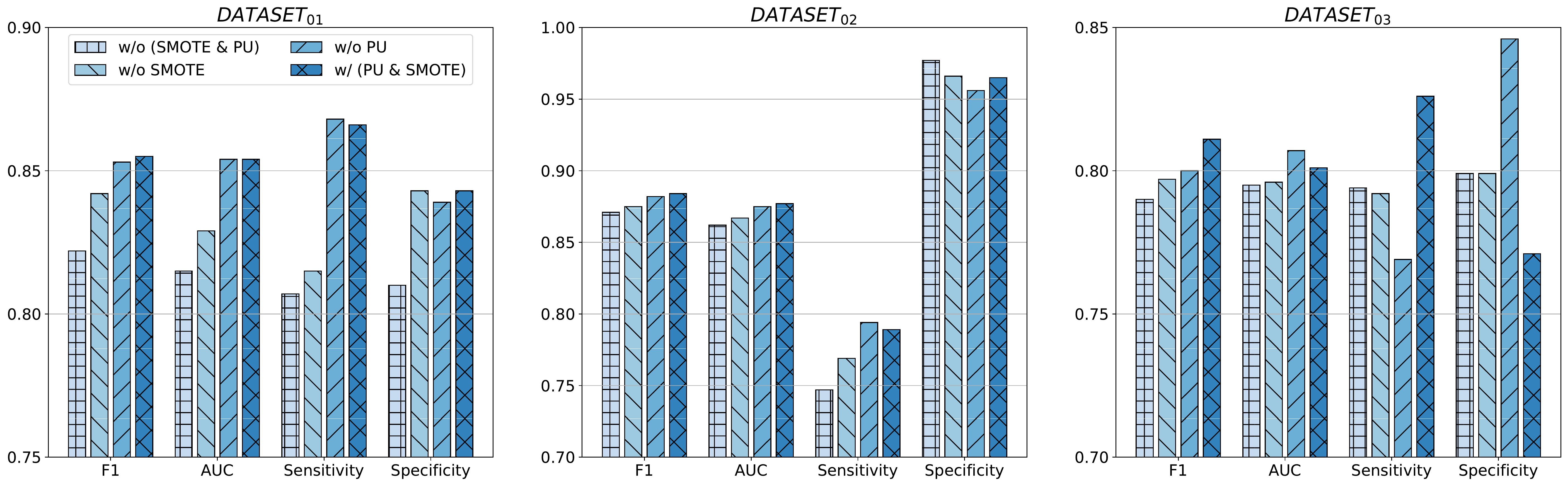}
    \caption{
        The bar plots of the effectiveness of PU and SMOTE with four different metrics. The comparison consisted of w/o (SMOTE \& PU), w/o SMOTE,  w/o PU, and w/ (PU \& SMOTE).
    }
    \label{figure_05_barplot}
\end{figure}

In Figure~\ref{figure_05_barplot}, the experimental results of the F1-score, AUC, sensitivity (recall; true positive rate), and specificity (true negative rate). First, the ablation study w/o (SMOTE \& PU) is conducted to verify whether there are solutions to deal with the imbalance label problem and uncertainty of the unlabeled dataset. As result, the metrics of this ablation study score the lowest among other ablation studies in all four measures. Second, the ablation study only w/o SMOTE (only with PU) is conducted to see the effect of the balancing process of the proposed method. We can verify that using GraphSMOTE \cite{zhao2021graphsmote} and its modification, the method can better handle the nature of label imbalance in chargeback fraud detection, scoring higher F1 score and AUC respectively. Third, the ablation study w/o PU (only with SMOTE) proves that it is proper to use PU loss to consider users who have not yet canceled payment as unlabeled. The users may cancel their payments after the inference of whether the users are chargeback fraud or not. As the users are considered unlabeled, training our proposed method with PU loss performs at least equal or better F1 score and AUC.

\section{Conclusion \& Feature Work}



We propose a novel chargeback fraud detection model, \textit{PU GNN}, for play-to-earn (P2E) MMORPGs. The proposed model leverages both players' in-game behavior log sequences and P2E token transaction networks. The model adopts an attention mechanism and graph attention networks to retrieve high-dimensional representation for the players. With the positive and unlabeled (PU) learning setup, the model is able to jointly learn positive (chargeback fraud) and unlabeled labels. The conducted experiments on three real-world datasets showed the proposed model outperforms other previously presented methods.

We believe there is still room for improvement. The time-related temporal features such as collusive work times and the time interval between the payment and its cancellation by the chargeback frauds are not yet considered which could be handled by adopting hazard function from survival analysis as features to be concerned. Another important aspect of graph-based fraud detection is handling various node types and edge types by extending homogeneous graph learning to heterogeneous graph learning. Early detection of chargeback fraud is also an important topic to consider to prevent and minimize losses. Therefore, we will further study the depth of these points to improve the performance of our proposed method.


\subsubsection*{Acknowledgements.} This research was supported by the Abnormal User Information Team, AI Center, Netmarble Corp.

\bibliographystyle{splncs04}
\bibliography{references}

\end{document}